\documentclass[10pt,twocolumn,letterpaper]{article}

\usepackage[pagenumbers]{cvpr} 

\usepackage{graphicx}
\usepackage{amsmath}
\usepackage{amssymb}
\usepackage{booktabs}
\usepackage{svg}
\usepackage{float}
\usepackage{pifont}
\usepackage{diagbox}
\usepackage[pagebackref,breaklinks,colorlinks]{hyperref}

\usepackage[capitalize]{cleveref}
\crefname{section}{Sec.}{Secs.}
\Crefname{section}{Section}{Sections}
\Crefname{table}{Table}{Tables}
\crefname{table}{Tab.}{Tabs.}

\def\cvprPaperID{*****} 
\def\confName{CVPR}
\def\confYear{2022}

\begin{document}

\title{Rink-Agnostic Hockey Rink Registration}

\author{Jia Cheng Shang\\
University of Waterloo\\
Waterloo, Ontario, Canada\\
{\tt\small jcshang@uwaterloo.ca}
\and
Yuhao Chen\\
University of Waterloo\\
Waterloo, Ontario, Canada\\
{\tt\small yuhao.chen1@uwaterloo.ca}
\and
Mohammad Javad Shafiee\\
University of Waterloo\\
Waterloo, Ontario, Canada\\
{\tt\small mjshafiee@uwaterloo.ca}
\and
David A. Clausi\\
University of Waterloo\\
Waterloo, Ontario, Canada\\
{\tt\small dclausi@uwaterloo.ca}
}

\maketitle

\begin{abstract}
Hockey rink registration is a useful tool for aiding and automating sports analysis. When combined with player tracking, it can provide location information of players on the rink by estimating a homography matrix that can warp broadcast video frames onto an overhead template of the rink, or vice versa. However, most existing techniques require accurate ground truth information, which can take many hours to annotate, and only work on the trained rink types. In this paper, we propose a generalized rink registration pipeline that, once trained, can be applied to both seen and unseen rink types with only an overhead rink template and the video frame as inputs. 
Our pipeline uses domain adaptation techniques, semi-supervised learning, and synthetic data during training to achieve this ability and overcome the lack of non-NHL training data.
The proposed method is evaluated on both NHL (source) and non-NHL (target) rink data and the results demonstrate that our approach can generalize to non-NHL rinks, while maintaining competitive performance on NHL rinks.
\end{abstract}

\section{Introduction}
\begin{figure}[h]
	\begin{center}
		\includegraphics[width=0.45\textwidth]{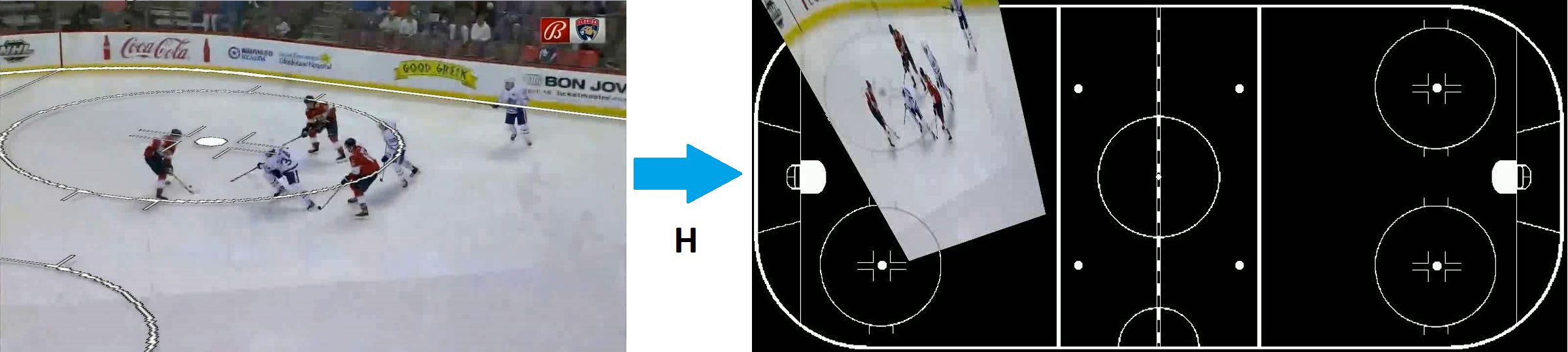}
		
	\end{center}
	\caption{Example of warping a video frame onto the overhead rink template (and vice versa) using homography.}
	\label{fig1}
\end{figure}

Rink registration plays a crucial role in automatic hockey game analysis. Rink registration is the process of mapping video frame pixels onto an overhead view of the rink template in order to determine the locations of everything on the ice. This location information of players is necessary for many types of further hockey analysis, such as interactions between players and determining better scoring opportunities. An example of how rink registration works can be seen in Figure \ref{fig1}.

\begin{figure}[h]
	\begin{center}
		\includegraphics[width=0.45\textwidth]{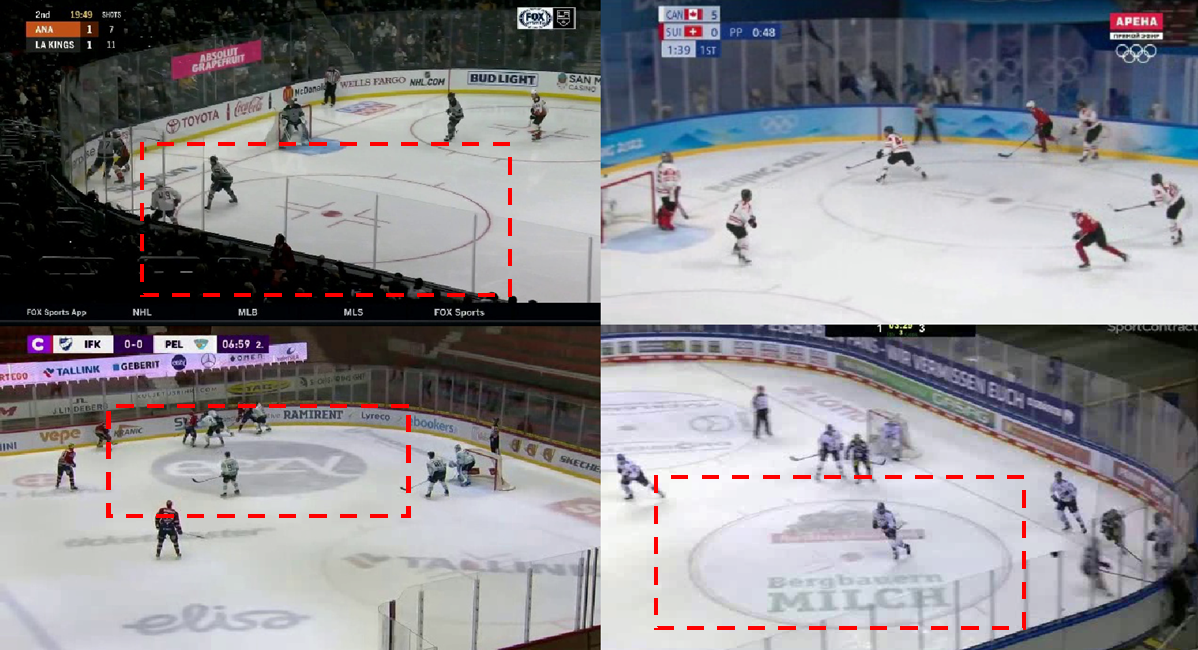}
		
	\end{center}
	\caption{Examples of different rinks. On top of the differences between rink shape and feature positioning, there are also differences in color, advertising frequency, and how face-off circles were filled. Face-off circle differences are highlighted using dashed boxes}
	\label{fig_rinks}
\end{figure}

Most existing rink registration systems focus on NHL rinks, which have a strict standardization system \cite{Weiner_2009} \cite{shi2022self} \cite{jiang2020optimizing} \cite{nie2021robust}. This means that each rink is the same size, with the same positions for rink features such as faceoff circles, blue lines, and goal lines. However, non-NHL rinks also exist. For example, many European rinks follow the International Ice Hockey Federation (IIHF)/Olympic hockey rink format, which is wider than the NHL standard \cite{IIHF}. This standardization is not as strict, resulting in varying rink sizes and feature location changes in different rinks. For example, some arenas in Finland have sizes that fall between IIHF and NHL sizes \cite{Formánek}. Also, minor leagues and recreational rinks may not follow standards as strictly, resulting in more differences. Examples of different rinks can be seen in Figure \ref{fig_rinks}. 

In these situations, rink registration systems trained on NHL data do not perform well, often resulting in incorrect warps in our tests. They would need new ground truth homography data on these new rinks in order to function correctly, which is a costly and time-consuming task. Furthermore, different rink setups and sizes would need different trained models. Thus, existing models lack generalizability for rink setups, and this is difficult to resolve without a large quantity of data from a variety of rinks.

We propose a  novel pipeline with three main modules (models) to resolve the aforementioned issues. The first model performs semantic segmentation on the input image to produce a segmentation map. The following two models estimate and refine a homography estimation based on the segmentation map and the corresponding rink template. To address the lack of data for non-NHL rinks, we implement domain adaptation techniques, use improved augmentations, and use synthetic data to simulate different possible rinks.

To the best of our knowledge, this pipeline is the first system designed for sports rink registration that is able to work on a variety of rink types, making it rink-agnostic. It is able to estimate homography for multiple rink types with competitive accuracy, despite only having labelled data for a single rink type.

\section{Related Works}
\label{sec:related_works}

\subsection{Homography Estimation}
Homography estimation is an aspect of image processing where one plane is warped onto another plane. Traditional homography estimation techniques involve identifying feature pairs from image pairs using methods such as SIFT \cite{lowe2004distinctive} and ORB \cite{rublee2011orb}, before being used in systems such as RANSAC and Direct Linear Transform (DLT) to calculate the homography matrix \cite{fischler1981random} \cite{hartley2003multiple}. 

DeTone {\it et al.} \cite{detone2016deep} were one of the first to estimate homography via deep learning, estimating the location of four corners of one image in the image space of the other. These sets of point estimates can then be converted into homography via DLT \cite{detone2016deep}, as matching 4 (x,y) pairs is enough to solve the 8 unknowns in the homography matrix.

Zhou and Li \cite{zhou2019stn} described how directly estimating homography parameters via deep learning was difficult due to the different scaling and distributions needed for each parameter. They normalized homography matrix parameters so that they have relatively similar distributions, making them more suitable for the loss functions used by deep learning models \cite{zhou2019stn}.

Since then, other models have been built for homography estimation, and these include various models specialized for sports field registration \cite{homayounfar2017sports}, \cite{chen2019sports}, \cite{sha2020end}, \cite{nie2021robust}, \cite{chu2022sports}, \cite{jiang2020optimizing}.

Our pipeline adopts the approaches from Jiang {\it et al.} \cite{jiang2020optimizing} and Shi {\it et al.} \cite{shi2022self}, where an estimate and refinement approach is taken for homography estimation, and self-supervised learning and synthetic data are used to improve the training process. It uses the 4-point approach popularized by \cite{detone2016deep}. 

\subsection{Semantic Segmentation}
Semantic segmentation involves classifying each pixel in an image into several provided categories. With the advent of deep learning, many models were developed to do this for fields such as autonomous driving and remote sensing. 

Long {\it et al.} ~\cite{long2015fully} popularized the use of fully convolutional networks for the purpose of semantic segmentation. Ronneberger {\it et al.} ~\cite{ronneberger2015u} designed the U-net model which builds upon the fully convolutional network by setting up a dedicated encoder and decoder structure with skip connections to improve the upscaling process.

The DeepLab series of models further build upon the U-net structure by adding various techniques such as Atrous Spatial Pyramid Pooling (ASPP) and image level pooling, to improve long range and global context information acquisition ~\cite{chen2017deeplab} ~\cite{chen2017rethinking} ~\cite{chen2018encoder}. Various vision transformer based approaches were also used for semantic segmentation, such as Segformer~\cite{xie2021segformer}, taking advantage of the improvements transformers provided to the field of image processing.

\subsection{Domain Adaptation for Semantic Segmentation}
Unsupervised domain adaptation (UDA) involves trying to bridge the domain gap caused by differences between the labelled training data (source domain) and unseen test data (target domain). UDA tries to mitigate this issue by training on both labelled source data and some unlabelled target data and using techniques to improve the model's performance in the target domain. Techniques such as maximum mean distances~\cite{long2015learning}, adversarial learning ~\cite{ganin2016domain}, and self-training ~\cite{tranheden2021dacs} have been used for deep learning in order to improve the model's ability to bridge the domain gap.

Self-training methods in particular seem to perform well for UDA in the field of semantic segmentation, with several recent works using it~\cite{hoyer2022daformer}~\cite{hoyer2022hrda}. DAformer by Hoyer et al. use a teacher-student approach for self-training, where a teacher model is gradually updated using the exponential mean average of the student weights and is used to produce pseudo-labels of the target data for the student to train on~\cite{hoyer2022daformer}. Masked Image Consistency (MIC) is another work that uses a similar approach that can be added on top of existing domain adaptation methods~\cite{hoyer2023mic}. It involves masking the target images fed into the student model in order to train it to learn contextual relations between different components in the target image. The loss is then computed between the predicted heatmap and a pseudo-label generated by the teacher model, which has access to the entire image.

We take inspiration from this field to improve performance on the target domain of non-NHL rinks, especially during the segmentation stage of the pipeline.

\subsection{Semantic Segmentation and Homography}
Some models have used homography to improve the results of semantic segmentation, especially in cases where the resulting segmentation is expected to follow a structure that is known beforehand. Examples can include organ semantic segmentation in biology, where the organ components have a roughly known structure, and this prior can be used to provide a better segmentation.

Lee {\it et al.} \cite{lee2019image} develop an Image-and-Spatial Transformer Network (ISTN), which consists of two components: an image transformer network (ITN) that generates a representation of two input images, and a spatial transformer network (STN) that is trained to find the affine transform needed to align the resulting feature representations together \cite{jaderberg2015spatial}. Sinclair {\it et al.} build upon this work in their Atlas-ISTN by setting the ITN to be a semantic segmentation network, and using the result of that in a STN to warp an ``atlas'' template to a proper orientation \cite{sinclair2022atlas}.

Our pipeline uses a similar approach of performing segmentation before estimating a warp matrix. However, the main goal is estimating the matrix used to warp the template, rather than getting the warped segmentation itself. Furthermore, we require a homography matrix rather than an affine matrix in order to map one plane onto another. We only have ground truth training data for a single source domain, and use UDA techniques to improve results on other rink types.

Other sports registration systems have used segmentation to extract feature information before further analysis \cite{homayounfar2017sports} \cite{zhang2021high}. However, none of them do so for the purpose of performing rink-agnostic homography.

\section{Methodology}
We propose an end-to-end system for rink-agnostic homography estimation. It takes in video frames and the overhead template of the rink as input, and outputs the homography needed to warp the template onto the frame. Our pipeline consists of 3 components:
\begin{enumerate}
    \itemsep0em
    \item A semantic segmentation model takes in the input video frame and outputs a semantic segmentation map.
    \item An initial homography estimator takes the segmentation map and the overhead rink template as input and outputs the homography needed to align the two together. This homography is then used to warp the overhead template and produce a warped template estimate.
    \item Finally, a refinement model takes the segmentation map and warped templates as input and produces a refinement homography to adjust the warped template estimate to be closer to the proper orientation seen in the segmentation map. This process can be iterated to further improve the homography.
\end{enumerate}

\begin{figure}
	\begin{center}
		\includegraphics[width=0.45\textwidth]{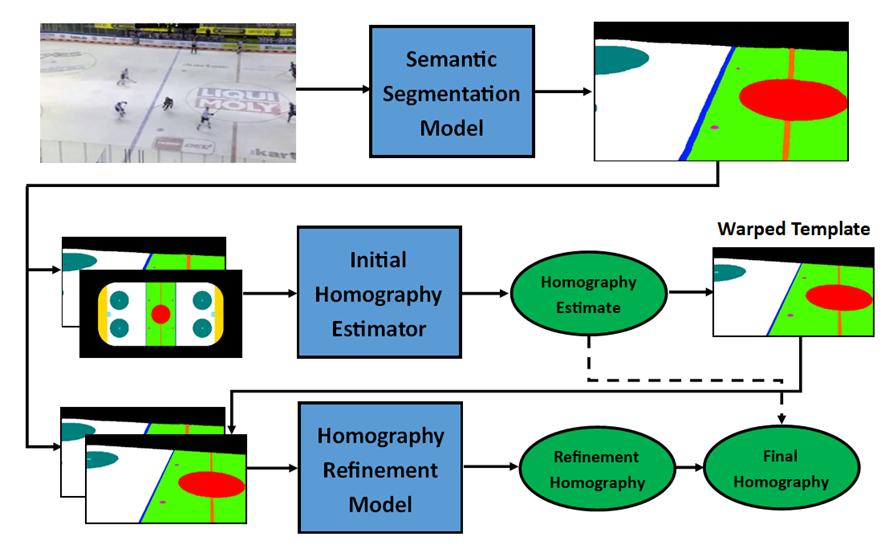}
		
	\end{center}
	\caption{Pipeline of the process during test time, showing the 3 major components. The inputs to the pipeline are the video frame fed to the segmentation model and the overhead template fed to the initial estimator. The iteration of the refinement model has been omitted for clarity.}
	\label{fig2}
\end{figure}

The overall pipeline and how the three components interact with each other can be seen in Figure~\ref{fig2}. However, we still have a lack of labelled training data for non-NHL rinks. To solve this issue, we use domain adaptation, augmentations, and synthetic data to train each component separately. This helps make the system more rink agnostic and helps overcome the lack of data for other domains. 

The segmentation module is trained via domain adaptation techniques on both labelled NHL and unlabelled non-NHL data. Augmentations such as logo augmentation are also added, to simulate differences in appearance between rinks and further improve generalizability. The other two modules are trained in a semi-supervised manner using synthetic data, in order to generalize them to different rink types.

\subsection{Semantic Segmentation Module}
\begin{figure}
	\begin{center}
		\includegraphics[width=0.45\textwidth]{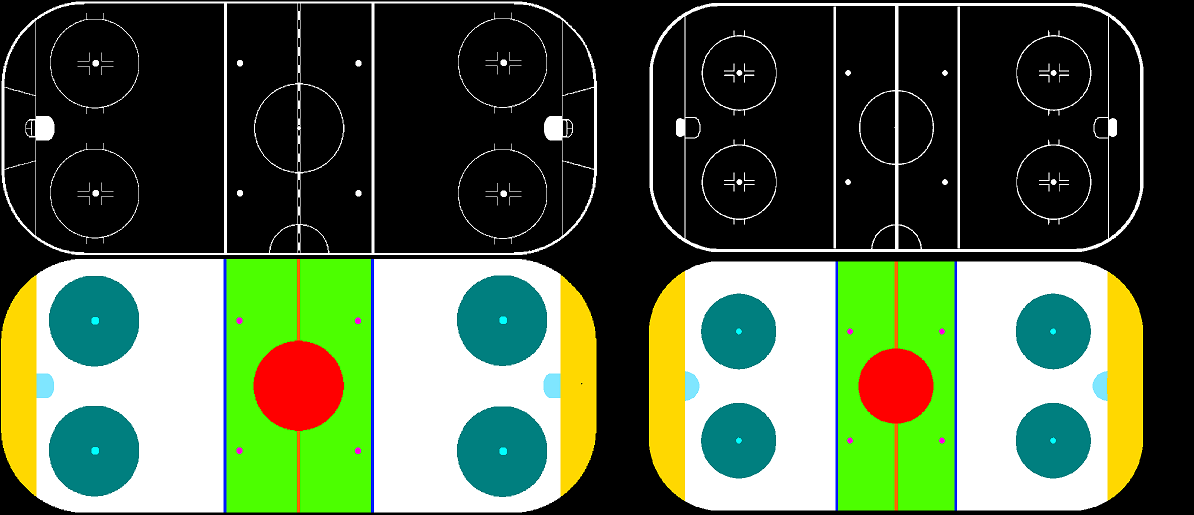}
		
	\end{center}
	\caption{The line and segmentation overhead templates used for NHL (left) and Olympic (right) rinks. Note that in reality, both rinks are the same lengthwise, and the Olympic rinks are wider than the NHL rinks. They were both scaled to fit the same template space for this analysis.}
	\label{fig_rink_templates}
\end{figure}

The semantic segmentation module is designed to identify the various rink features in broadcast video frames, regardless of the type of rink used. Different rinks such as NHL and Olympic rinks can have different structures, and there usually isn't a scaling or direct linear transformation that can warp the rinks to be the same form. These rinks are seen in Figure \ref{fig_rink_templates}. However, although the various features such as faceoff circles and blue lines may differ in size and positioning, they will still exist in all major rinks. This allows them to be used as classes for semantic segmentation regardless of which rink the image was taken from. 

In order to improve the model's ability to generalize on all rinks, we used heavy augmentation as well as domain adaptation techniques. On top of general augmentations such as Gaussian noise, color augmentation, shifts, tilts and zooms, we added copy-paste augmentation and logo augmentation. 

\subsubsection{Augmentations}

Copy-paste augmentation is based on the work of Ghiasi {\it et al.}~\cite{ghiasi2021simple}. However, their copy-paste system involved pasting instances from one image onto the other in order to improve the instance segmentation of items in different scenarios. In our case, we copy-paste players from other images in order to simulate the natural occlusion of rink features. This is used to improve the model's ability to segment rink features even when they are occluded. 

\begin{figure}
	\begin{center}
		\includegraphics[width=0.45\textwidth]{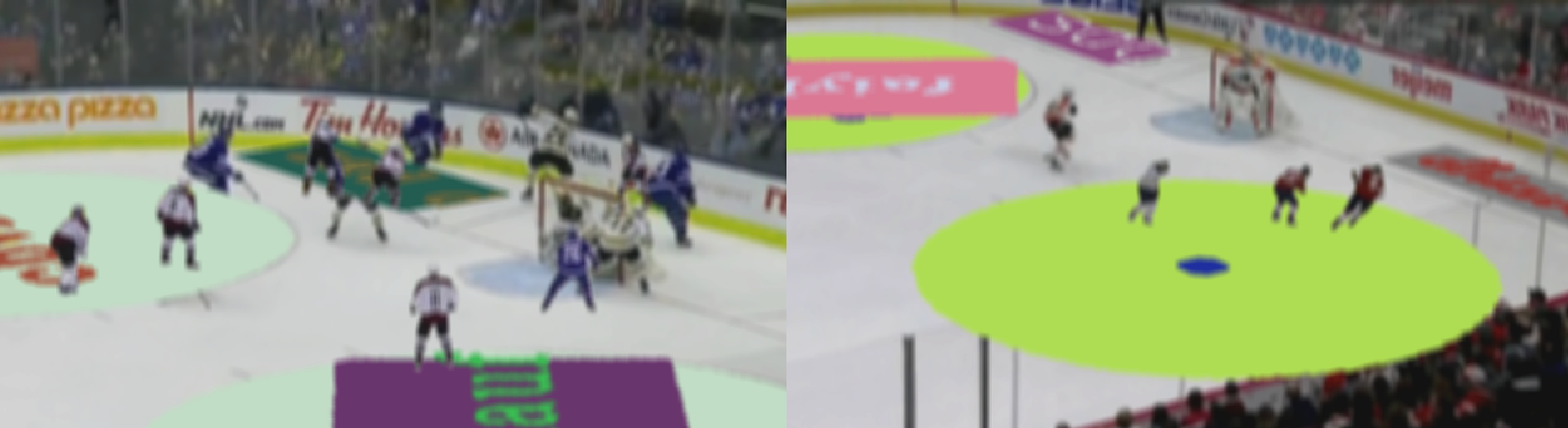}
		
	\end{center}
	\caption{Examples of logo augmentation, which sometimes added text, rectangles, and circle fillings in order to augment the existing dataset further.}
	\label{fig_logos}
\end{figure}

Logo augmentation is designed to simulate the random advertising and text that may appear on different rinks. Randomized text, rectangles, and circle fillings are added in areas with space that may have logos in some rinks. This is done to teach the network to ignore the effects of such advertising. Examples can be seen in Fig.~\ref{fig_logos}.

\subsubsection{Domain Adaptation}

We also use domain adaptation in order to improve the model's performance on Olympic rinks, where we do not have any ground truth segmentation training data. In particular, we adopt some methods described in MIC~\cite{hoyer2023mic}, to improve the model's ability to learn the context between different components in the target domain. This would allow us to use unlabelled non-NHL data during our training.

We primarily add the exponential moving average (EMA) teacher-student and input masking behavior to our pipeline, as described in \cite{hoyer2023mic}. The EMA teacher-student approach has been shown to improve results for semi-supervised training \cite{tarvainen2017mean} \cite{hoyer2022daformer}, and in domain adaptation self-training. In this case, the target domain of non-NHL rinks is unlabelled and pseudo-labels generated by the teacher are used instead. So during training, we have a student model that learns via loss functions, and a teacher model who's weights are altered over time based on the EMA of the student's weights over time. When training on the unlabelled target domain, the teacher has access to the unmasked image, and produces a pseudo-label. 

\begin{figure}
	\begin{center}
		\includegraphics[width=0.45\textwidth]{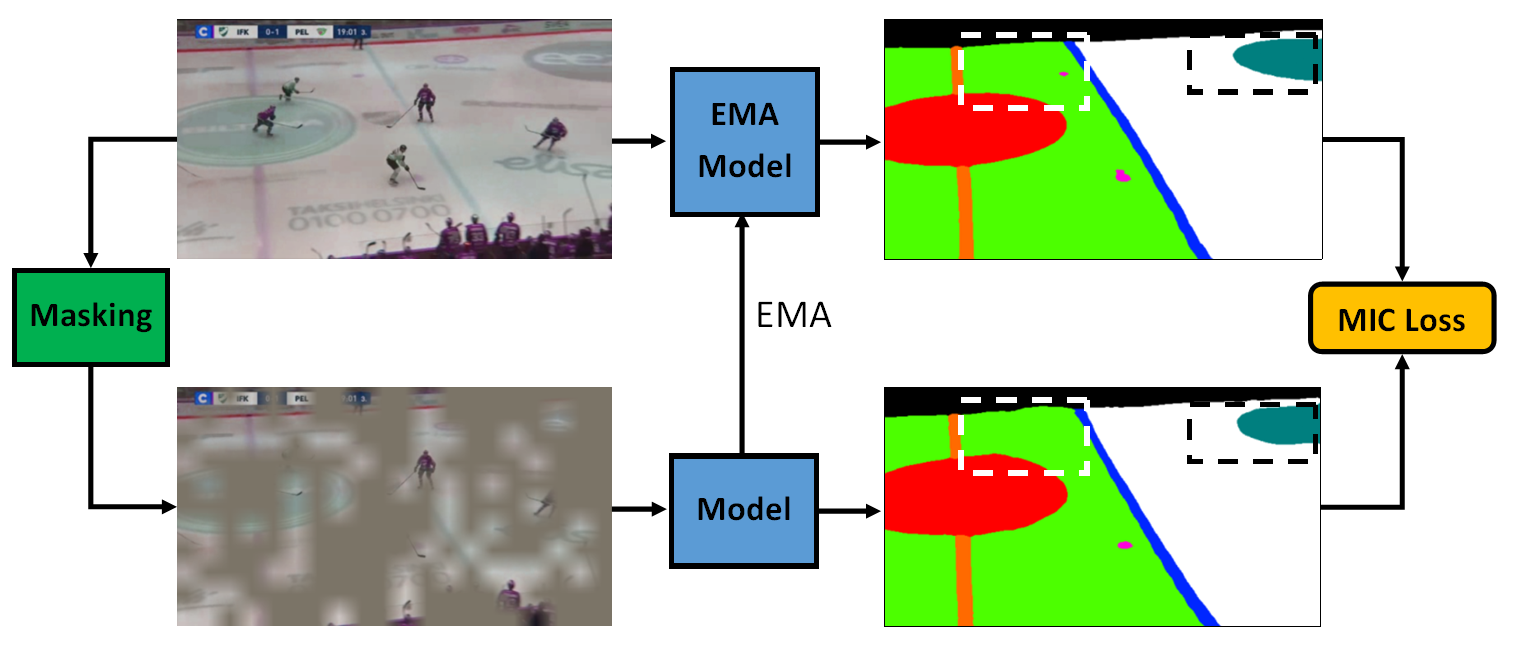}
		
	\end{center}
	\caption{Pipeline of MIC method from \cite{hoyer2023mic}. It promotes the model to learn contextual clues as it needs to identify the hidden areas based on information from other non-hidden areas. The dashed rectangles highlight some areas that the model needs to correct.}
	\label{fig_MIC}
\end{figure}

The student, however, only has access to the masked input and produces a segmentation mask which is compared against the pseudo-label with a segmentation loss. This loss is weighted by the confidence weighting of the pseudo-label (as pseudo-labels may not be precise), and used to update the student model. The teacher's weights are then updated in turn via the EMA equation, as seen in equation \ref{eqn_ema}, where \textit{t} denotes timestep, \textit{$\Phi$} denotes teacher weights, \textit{$\Theta$} denotes student weights, and \textit{$\alpha$} is a smoothing factor \cite{tarvainen2017mean}. The usage of MIC is seen in Figure ~\ref{fig_MIC}.

\begin{equation}\label{eqn_ema} \Phi_{t+1} \leftarrow \alpha\Phi_{t} + (1-\alpha)\Theta_{t} \end{equation}

We used a DeepLabV3+ model \cite{chen2018encoder} from the Segmentation Models PyTorch library \cite{Iakubovskii:2019} as the segmentation model in this case. We also used focal loss for the segmentation loss \cite{lin2017focal}, and AdamW optimizer.

\subsection{Homography Estimator Module}
The homography estimator module consists of a Resnet18-based regressor that estimates the normalized homography matrix, in a similar manner as \cite{zhou2019stn}. During the inference time, it takes the segmentation map output of the first module alongside an overhead template of the rink as input. It then produces an estimate of the homography needed to warp the overhead template to be aligned with the segmentation map (which makes it also aligned with the actual input frame if the segmentation map is accurate). 

\begin{figure}
	\begin{center}
		\includegraphics[width=0.45\textwidth]{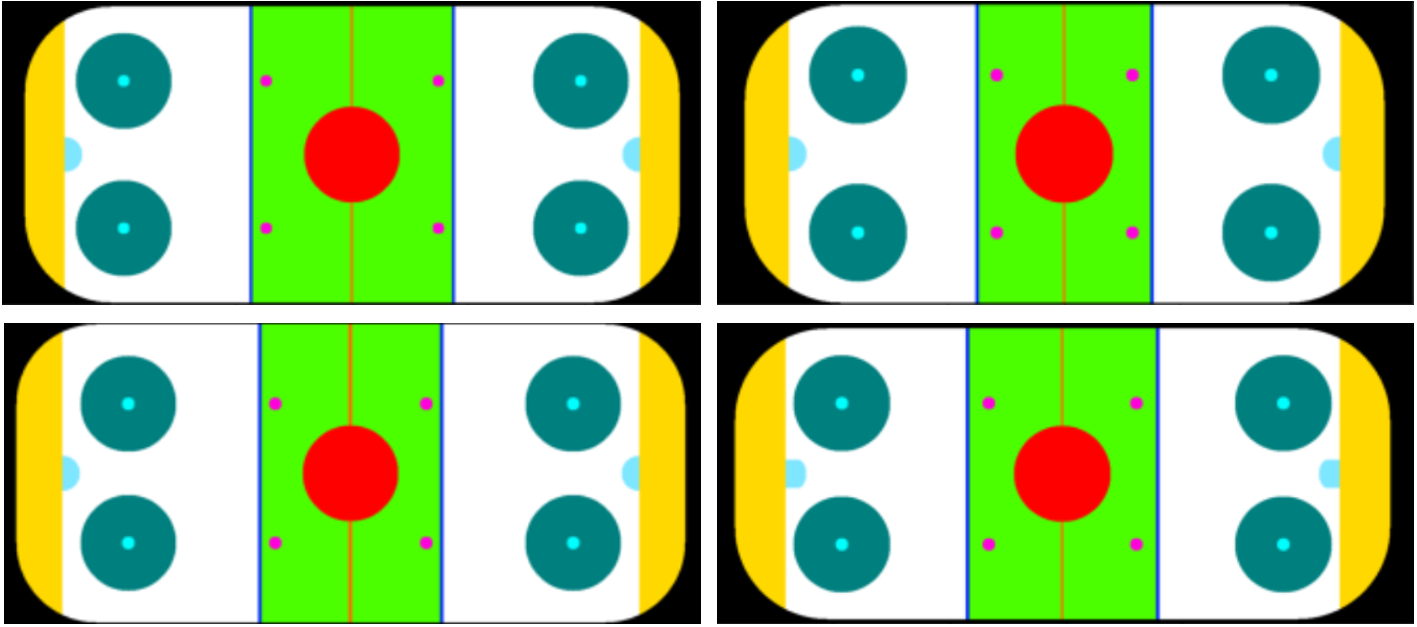}
		
	\end{center}
	\caption{Examples of randomly generated rinks. The feature types and rough positions were kept constant, while the sizes, scales, and more precise positioning was varied each time. Some differences include goal crease shape, wider rinks having more space between faceoff circles and edges, and the blue lines and goal lines being in shifted locations.}
	\label{fig_random_rinks}
\end{figure}

However, during training, we use synthetic data because we only had labelled training data for NHL rinks. In order to generalize well on all rink setups and sizes, we use synthetic rink generation to simulate different rink setups. This is done by altering various distances in the overhead template, such as the distance between faceoff circles and the goal line, or the distance between blue lines and the center line. Examples of these can be seen in Fig.~\ref{fig_random_rinks}. 200 synthetic rinks were generated for training.

During data generation, we choose from common pre-defined rinks such as NHL or Olympic rinks, or create our own randomly generated rink to serve as the initial overhead rink template. This would thus improve its accuracy on a wide variety of rinks, as the model would be trained on a wide variety of templates. 

\begin{figure}
	\begin{center}
		\includegraphics[width=0.45\textwidth]{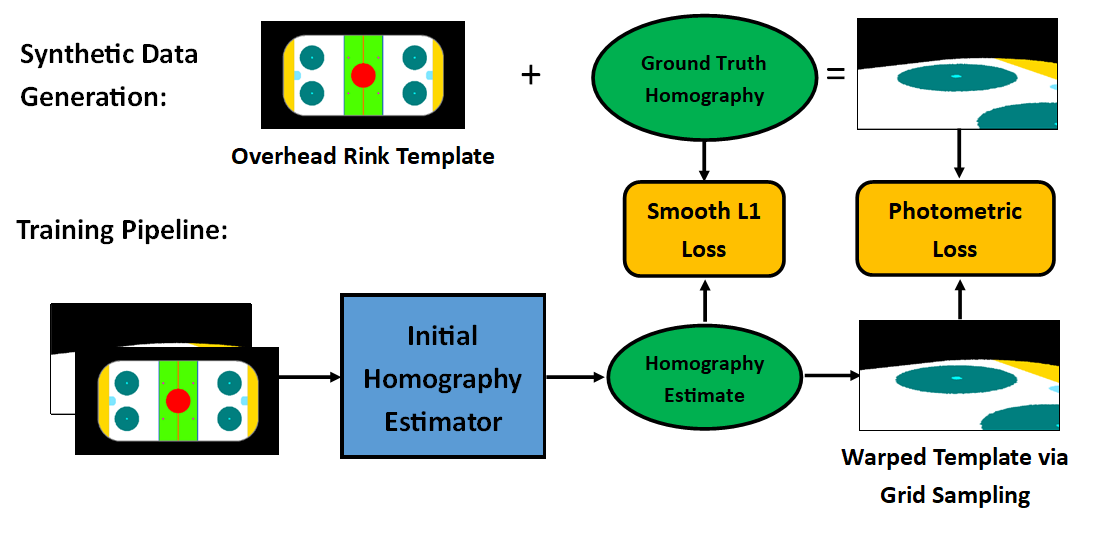}
		
	\end{center}
	\caption{Data generation and training process of the homography estimation module. The overhead template and warped template are used as input, and the resulting homography and warped output template are compared with the ground truth. Note the normalization and unnormalization of the homography is omitted in this image for clarity.}
	\label{fig_h_train}
\end{figure}

The next step in data generation involves acquiring a homography to warp the overhead rink to create a warped template. To do this, we use a ground truth homography matrix from the NHL dataset, and augment it with slight perturbations, zooms, and flips. The resulting warped template simulates what a segmentation mask input would look like, and is used as the synthetic data. This process can be seen in Fig.~\ref{fig_h_train}. We use ground truth homographies from the NHL training set to represent the range of homographies that correspond to broadcast video. The augmentations applied to the homography matrix help cover this expected range. It also covers potential differences in homography ranges that may occur when we use different templates, as the rink sizes can differ in those cases.

During training, the overhead rink template and warped template are fed as input to the initial homography estimator, which estimates the normalized homography. This homography is then used to warp the overhead template to produce a warped template output. This is done using grid-sampling, which preserves the gradient flow and allows the loss to be propagated back to the estimator model. The normalized homography estimate is compared with the ground truth homography via smoothed l1-loss, while the warped template output is compared with the original warped template via L1 loss. This process can be seen in Fig.~\ref{fig_h_train}. 

Note that the model only outputs a homography, so it cannot directly produce a copy of the warped template that was given as input. Thus, it needs to learn the homography required to warp the overhead template to the warped template input. During test time, we take the homography estimate and use it for the next module in the pipeline.

\subsection{Refinement Module}

The final module in the pipeline is the refinement model. During test time, its input consists of the segmentation mask from the first module alongside a warped template using the homography estimate from the second module. During training however, we once again leverage the use of synthetic data and semi-supervised learning to improve the model's performance on multiple rink types. 

\begin{figure}
	\begin{center}
		\includegraphics[width=0.40\textwidth]{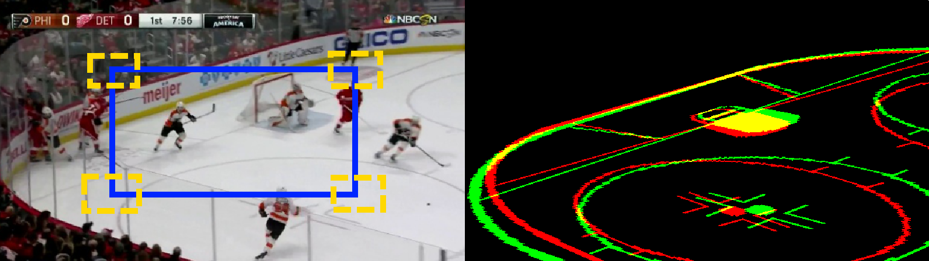}
		
	\end{center}
	\caption{Data generation for refinement. The blue rectangle represents an example initial four corners, and the yellow rectangle represent the possible perturbations for this example. The shift in homography can be seen on the right, with green being the original rink position and red being the perturbed version.}
	\label{fig_ref_data}
\end{figure}

For training data generation, we follow a similar scheme as Shi {\it et al} \cite{shi2022self}, where we take an existing ground truth homography and image, and augment them before feeding them into the model for training. This augmentation step involves selecting 4 random points in a rectangle on the image and perturbing them by a small amount. The previous and new positions of these points can then be used to produce a homography matrix, which is used to warp the image. The warped image and original image are then sent to the model during training, and it tries to calculate the homography needed to perform this warping process. In our case however, rather than using the video frame directly, our image consists of the overhead template warped by an existing ground truth homography. This homography is augmented before use, and is used to represent examples of rink orientations as viewed by the camera. This process is visualized in Fig.~\ref{fig_ref_data}. 

\begin{figure}
	\begin{center}
		\includegraphics[width=0.45\textwidth]{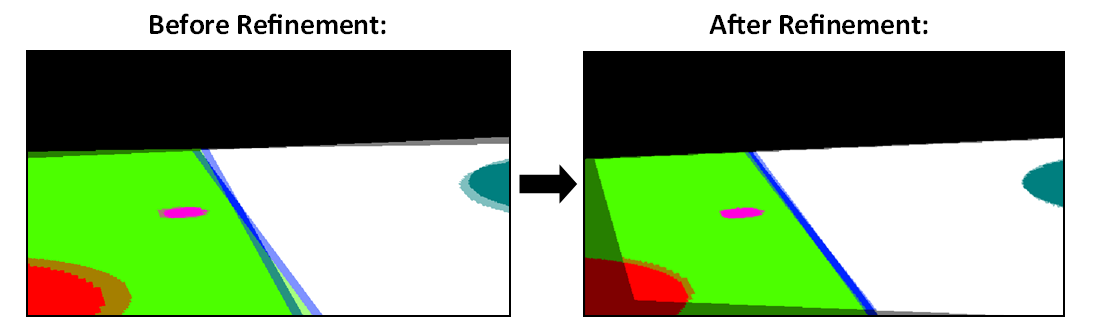}
		
	\end{center}
	\caption{Example of refinement. The left side shows the two input images overlaid on each other, and the right side shows the alignment that can occur after the refinement matrix is calculated.}
	\label{fig_ref_example}
\end{figure}

The refinement model is a Resnet18-based regressor, and uses the four-point approach to estimate homography, where it estimates the locations of 4 points from one image in the image space of the other. These sets of points can then be converted into a refinement homography via DLT~\cite{hartley2003multiple}, and will represent the warp needed to align the two input images. An example of pre-refinement inputs and a resulting refinement can be seen in Fig.~\ref{fig_ref_example}.

\begin{figure}
	\begin{center}
		\includegraphics[width=0.40\textwidth]{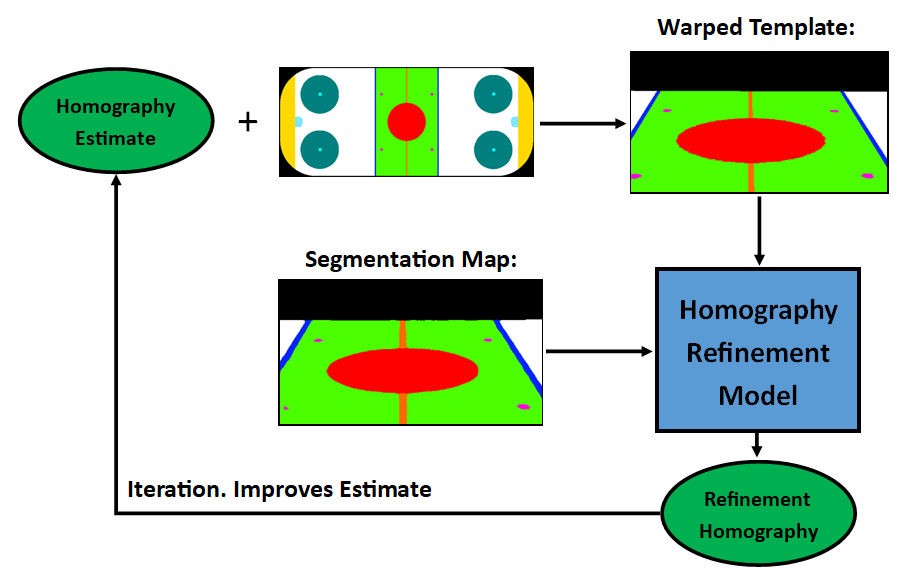}
		
	\end{center}
	\caption{Refinement iteration during testing. The resulting refinement matrix can be combined with the initial homography estimate to create a better warped template, which is fed into the refinement model again.}
	\label{fig_ref_iter}
\end{figure}

During test time, the refinement process can be iterated to further improve the homography refinement. The refinement homography can be combined with the initial estimate to produce a better estimate. This estimate is then used to warp the overhead template to produce a better warped template, which is fed back as input alongside the segmentation map. The refinement model performs this warp estimation process repeatedly, improving the alignment each time. In practice however, the alignment is only improved for the first few times, as small misalignments may not be aligned properly. Thus, we restrict the iteration at test time to 3 iterations, as we found not much improvement beyond that. This process is visualized in Fig.~\ref{fig_ref_iter}.

\section{Results and Discussion}

We first present experimental results for each component in the pipeline, and then describe the results for the overall pipeline.

\subsection{Segmentation Module}
The segmentation module was trained to predict 11 different classes of pixels from input images, including general areas such as background and defense zones to more specific features such as center line and goals. 

The copy-paste and logo augmentation did not affect the overall results on the source domain overall by much, as seen in Table \ref{table:seg_overall_iou}. Qualitatively however, they were able to improve the model's ability to identify parts occluded by players. Some features such as face-off spots can be occluded completely by players for example. Therefore, these augmentations help the model to learn to segment such features even when they are occluded by people, as the model would be trained on more examples of such cases.

\begin{table}
\small
\centering
  \caption{Overall average Intersection over Union (IOU) results from the validation set of NHL games. Although the augmentation and domain adaptation (DA) improvements did not affect the overall numbers much, they produce qualitative improvements on the target dataset results.}
  \label{table:seg_overall_iou}
  \begin{tabular}{ccl}
    \toprule
    Model & Overall IOU \\
    \midrule
    NHL-Only Model & 78.3\% \\
    Model with Augmentations & 78.6\% \\
    Model with Augmentations and DA & 78.5\% \\
  \bottomrule
\end{tabular}
\end{table}

We perform a sanity check by comparing an NHL-only trained model with the domain adaptation trained model, to ensure the accuracy did not drop on a validation set of NHL rinks. This can be seen in Table \ref{table:seg_overall_iou}. 

Results on NHL rinks can be seen in Table~\ref{table:seg_val}. Accuracy for these segmentations are measured via intersection over union (IOU), a common metric for this type of task. For the source domain validation set, we can see how the results are good for classes that cover areas, but have more errors in classes that represent lines or spots. This is partially because lines and spots are more likely to be obscured by players or the boards at the bottom of the rink, and any small deviation in prediction can cause a large IOU drop. 

\begin{table}
\small
\centering
\begin{center}
  \caption{Intersection over Union (IOU) results per class for segmentation models trained on source domain vs both domain. These results are the validation results from a set of held-out data on other NHL rinks and matches (source domain).}
  \label{table:seg_val}
  \begin{tabular}{cccl}
    \toprule
    Class & Single-Domain & Domain Adaptation\\
    \midrule
    Background & \textbf{97.1\%} & 97.0\% \\
    Behind Goal & \textbf{87.0\%} & 86.2\% \\
    Blue lines & 45.4\% & \textbf{51.3\%} \\
    Center Face-off Circle & \textbf{95.2\%} & 95.0\% \\
    Center Line & \textbf{62.3\%} & 60.6\% \\
    Outer Face-off Circles & 94.4\% & 94.4\% \\
    Outer Face-off Spots & \textbf{61.6\%} & 61.0\% \\
    Goal Creases & \textbf{81.8\%} & 81.5\% \\
    Neutral Zone & \textbf{94.9\%} & 94.4\% \\
    Inner Face-off Spots & 46.8\% & \textbf{47.2\%} \\
    Defense Zones & 94.8\% & \textbf{94.9\%} \\
    \hline
    Overall Average & 78.3\% & \textbf{78.5\%} \\
  \bottomrule
  \end{tabular}
\end{center}
\end{table}

The semantic segmentation module was also tested on various unlabelled target domain data, such as Olympic/European rinks. Although no labels, and thus no quantitative results, are available, qualitative analysis can still be done, where the predicted segmentation map is compared with the original image, to see if the components line up. 

\begin{figure}
  \centering
  \centering
  \begin{subfigure}[b]{0.22\textwidth}
     \centering
     \includegraphics[width=\textwidth]{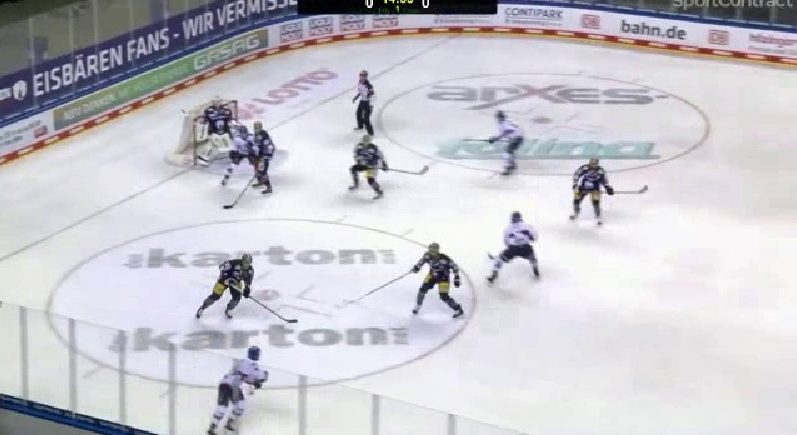}
     \caption{Input Image 1}
 \end{subfigure}
 \begin{subfigure}[b]{0.22\textwidth}
     \centering
     \includegraphics[width=\textwidth]{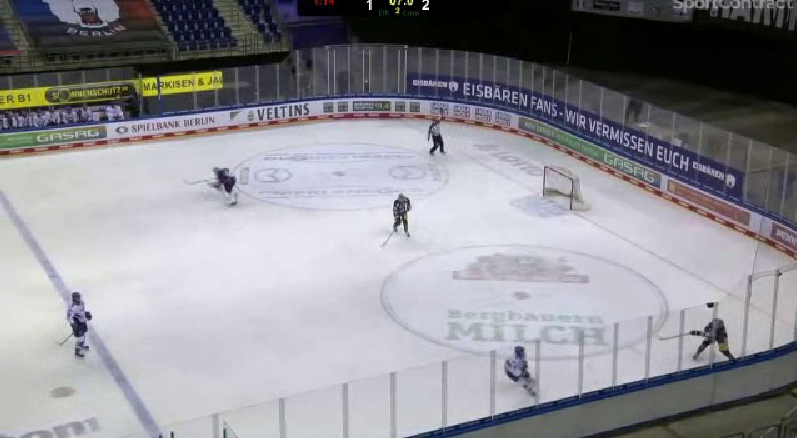}
     \caption{Input Image 2}
 \end{subfigure}
 \begin{subfigure}[b]{0.22\textwidth}
     \centering
     \includegraphics[width=\textwidth]{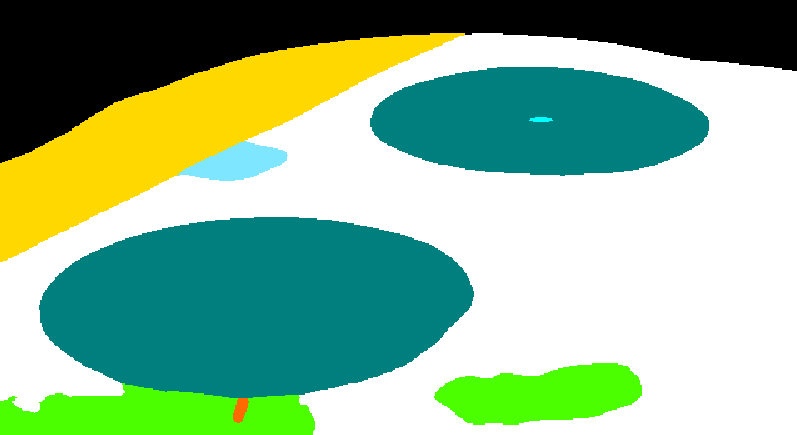}
     \caption{Base Segmentation}
 \end{subfigure}
 \begin{subfigure}[b]{0.22\textwidth}
     \centering
     \includegraphics[width=\textwidth]{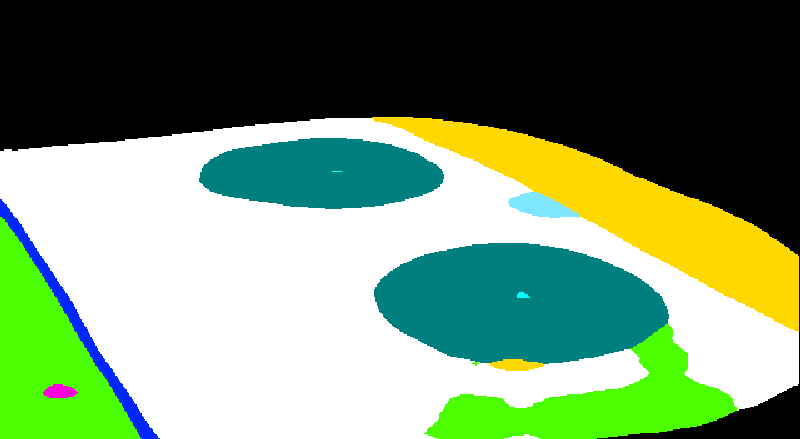}
     \caption{Base Segmentation}
 \end{subfigure}
 \begin{subfigure}[b]{0.22\textwidth}
     \centering
     \includegraphics[width=\textwidth]{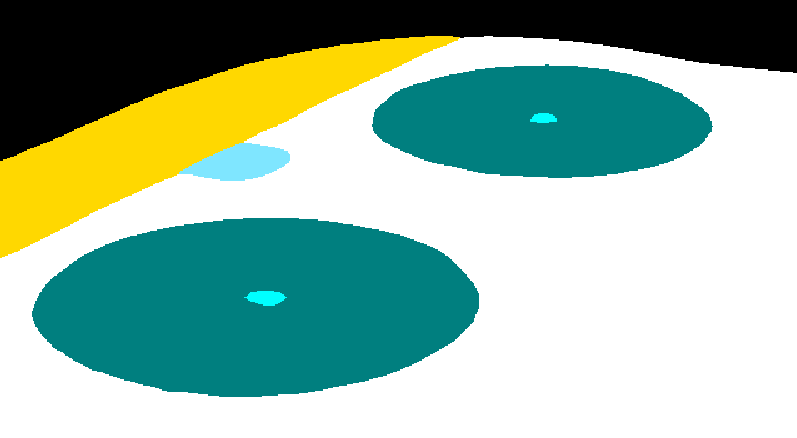}
     \caption{Improved Segmentation}
 \end{subfigure}
 \begin{subfigure}[b]{0.22\textwidth}
     \centering
     \includegraphics[width=\textwidth]{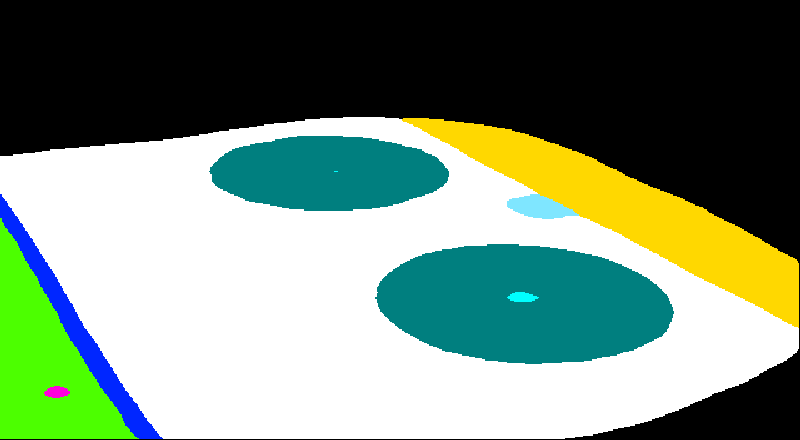}
     \caption{Improved Segmentation}
 \end{subfigure}
\caption{Examples of Olympic style rink images and corresponding predicted segmentation maps from a NHL-only model (c,d) and an improved model with logo augmentation and DA (e,f). Domain adaptation and logo augmentation have improved the generalization capabilities of the model, allowing it to segment this new rink better than the no DA model.}
  \label{fig:seg_val_olympic}
\end{figure}

Examples of this can be seen in Fig.~\ref{fig:seg_val_olympic}, using results on the Olympic and European validation set, which has different Olympic-style rinks not seen by either model during training. The results of the model trained with domain adaptation and our copy-paste and logo augmentations were noticeably better. In particular, cases of major misclassified regions and missing regions that were present in the predictions from the base segmentation model were fixed in the domain adaptation trained model. Thus, the domain adaptation model resulted in better results on the target domain of non-NHL rinks, when compared with the NHL-only trained model. This shows that even with heavy augmentation, the changes between NHL and non-NHL rinks can still be quite large, resulting in a domain gap that needs to be bridged in another way.  

\subsection{Homography Estimator Module}
The homography estimation module is designed to roughly estimate the homography needed to warp the rink template onto an input image (or vice versa, as warping in the reverse direction just requires inverting the homography matrix). In order to compare homography results for homography estimation, we use IOU\textsubscript{part}, where only the portion of the rink template that would have been in the image is considered. We use IOU\textsubscript{part} because the ground truth data collected was primarily done with just the visible portion in mind, and thus the ground truth for IOU\textsubscript{whole} may not have been accurate. The image is warped using both the predicted homography and the ground truth homography, and the resulting intersection and union are calculated. 

As with the semantic segmentation model, we compared a model trained solely on the source domain NHL rink with another model trained on multiple rink types and randomly generated rinks. This helps determine the viability of a rink-agnostic homography estimator. The source-domain trained initial estimator model performed 1.6\% better than the multi-rink trained model on the NHL validation dataset, as seen in Table~\ref{table:h_val}. 

However, the multi-rink trained model is still competitive and has the added benefit of working for multiple types of rinks, whereas the NHL-only model results were often off. The refinement module, later on, is used to improve the accuracy of the warps.

\begin{table}
\small
\centering
  \caption{Average Intersection over Union (IOU) results using ground truth homographies from the validation set of NHL games. Different stages in the pipelines are compared, such as Initial Estimator Model (IEM), Refinement Model (RM), and Iterative Refinement (IR). The multi-rink model with iterative refinement achieves similar accuracy as the NHL-Only pipeline on our data. However, it has the added benefit of working on non-NHL rinks as well.}
  \label{table:h_val}
  \begin{tabular}{ccl}
    \toprule
    Pipeline & IOU\textsubscript{part} \\
    \midrule
    NHL-Only Baseline IEM & 96.0\% \\
    Rink-Agnostic IEM & 94.4\% \\
    \midrule
    NHL-Only Baseline IEM + RM + IR \cite{shi2022self} & 96.9\% \\
    Rink-Agnostic IEM + RM & 96.7\% \\
    Rink-Agnostic IEM + RM + IR & 96.9\% \\
  \bottomrule
\end{tabular}
\end{table}

\subsection{Refinement Module}
The refinement model is the last component of the pipeline. It is designed to determine small homography differences between the segmentation mask output of the first model and the warped template created using the homography estimate of the second model. The refinement model must accurately calculate the homography needed to align the two inputs, and is trained on multiple fixed rinks and randomly generated rinks. 

The refinement process results on the synthetic data used in validation have an accuracy of approximately 98\%  IOU\textsubscript{part}.

\subsection{Overall Pipeline}
The results of the overall pipeline were analyzed to determine how well this system works on both NHL and non-NHL data. We use a model based on \cite{shi2022self} as the baseline, which was replicated because the source code, original model, and data were unavailable to the public. Using the source domain NHL validation set, it was found that the results of our pipeline are roughly on par with that of the baseline, as seen in Table~\ref{table:h_val}.

\begin{figure}[!h]
	\begin{center}
		\includegraphics[width=0.45\textwidth]{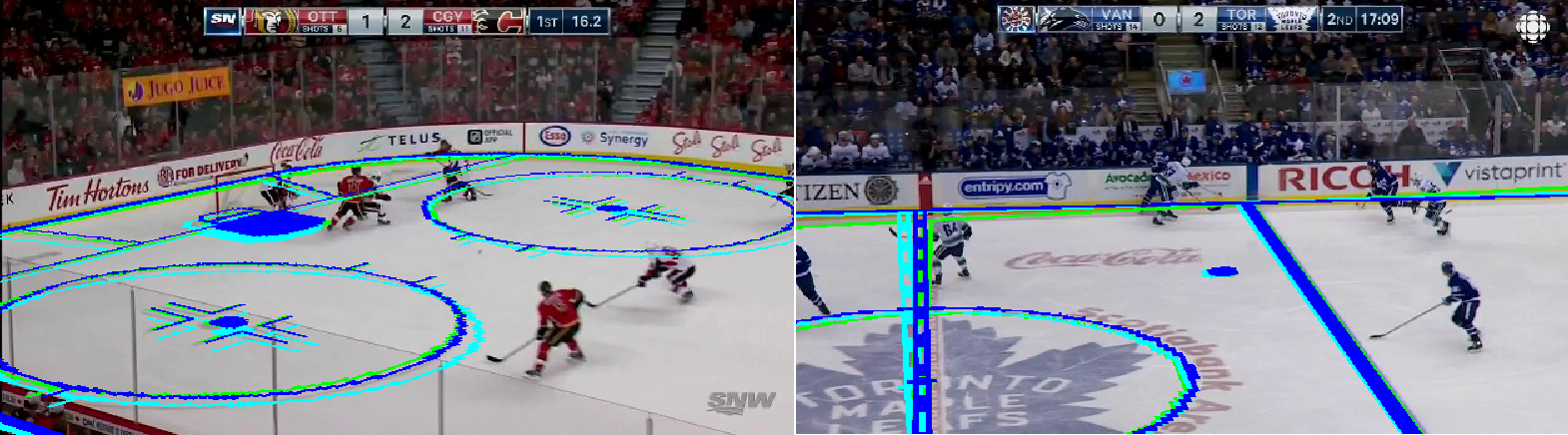}
		
	\end{center}
	\caption{Example results on NHL validation data. Green is ground truth, light blue is initial homography estimate, and dark blue is the final homography after refinement.}
	\label{fig_h_compare_nhl}
\end{figure}

The visual results for this approach on the NHL validation data can be seen in Fig.~\ref{fig_h_compare_nhl}. As seen in the images, the pipeline can warp the template to be closely aligned to the markings on the rink. 

Adding iteration to refinement improves the accuracy by a small amount as the refinement module can make further adjustments to the estimated homography after applying the previous refinement to the estimate. We compared single round refinement vs iterative refinement and saw the overall results on the NHL data improved by a small amount with iteration, as seen in Table~\ref{table:ref_iter}. Further adding iterations to refinement beyond 3 iterations did not increase the result by a meaningful amount. 

\begin{table}
\small
\centering
  \caption{Average Intersection over Union (IOU) results using ground truth homographies from the validation set of NHL games. The first and second iterations of the refinement module have the largest effect, while later iterations do not have much effect. We stop at 3 iterations as results do not change much after.}
  \label{table:ref_iter}
  \begin{tabular}{ccl}
    \toprule
    Pipeline With Different Refinement Iterations & IOU\textsubscript{part} \\
    \midrule
    No Refinement & 94.4\% \\
    Refinement with 1 Iteration & 96.7\% \\
    Refinement with 2 Iteration & 96.9\% \\
    Refinement with 3 Iteration & 96.9\% \\
  \bottomrule
\end{tabular}
\end{table}

\begin{figure}
	\begin{center}
		\includegraphics[width=0.41\textwidth]{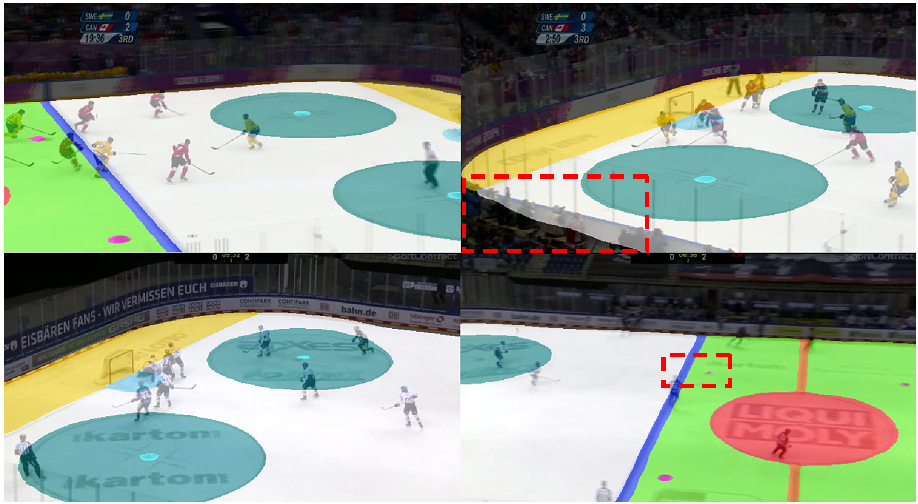}
		
	\end{center}
	\caption{Examples of segmentation on Olympic rinks. The left two show cases where the segmentation is cleaner, whereas the right two shows some more obvious defects. These include missing spots or over/underflowing edges (marked in red boxes).}
	\label{fig_seg_examples}
\end{figure}

On Olympic data, the resulting warps are usually close, but there are sometimes qualitative issues where the alignment is off. One potential cause for this involves problems during the segmentation stage, where the segmentation maps aren't fully accurate. Sometimes, regions such as face-off circles or the bottom edge of the rink may not be segmented accurately and can be off from their true location, as seen in Fig.~\ref{fig_seg_examples}. 

Overall results on 2 types of non-NHL rinks (Olympic 2014 and Berlin Mercedes-Benz Arena rink) can be seen in Table \ref{table:non_nhl}. Here we see that the NHL-only model results aren't as good, even after scaling is provided to make the template closer to the NHL template. The rink-agnostic model is more robust to arena template changes as it performs better than the baseline with these non-NHL rinks. It is more robust to rink appearance differences as well, since the performance of the NHL-model dropped dramatically on the Berlin arena, whereas the rink-agnostic model is not affected much. 

\begin{table}
\small
\centering
  \caption{Average IOU\textsubscript{part} results on a couple of non-NHL rinks. The rink-agnostic model outperforms the NHL-only baseline, especially in the Berlin arena where there are more differences in rink appearance.}
  \label{table:non_nhl}
  \begin{tabular}{cccl}
    \toprule
    Pipeline & Berlin Arena & Olympic 2014 Arena \\
    \midrule
    NHL-only Baseline & 87.7\% & 96.2\% \\
    Rink-Agnostic Model & \textbf{96.5\%} & \textbf{97.3\%} \\
  \bottomrule
\end{tabular}
\end{table}

\begin{figure}
	\begin{center}
		\includegraphics[width=0.41\textwidth]{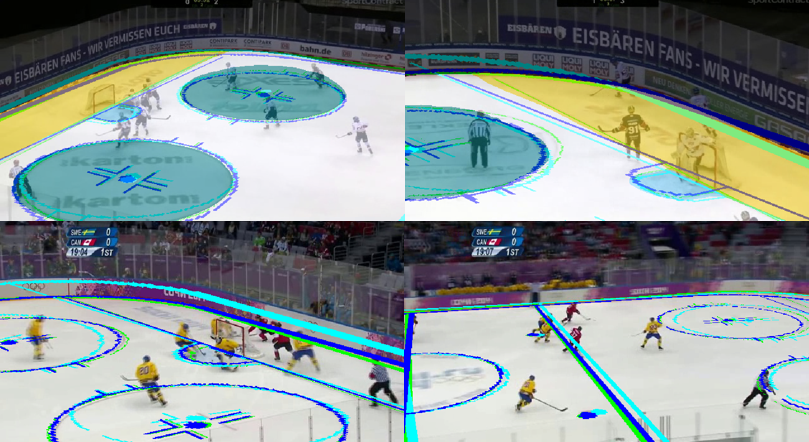}
		
	\end{center}
	\caption{Example results on Olympic validation data. Light blue is initial homography estimate, dark blue is the final homography after refinement, and green indicates ground truth. These cases show that although the alignment can be quite close usually, sometimes the alignment can still be off even after refinement for the Olympic rinks, likely due to the difficulty in segmenting these rinks.}
	\label{fig_compare_olympic}
\end{figure}

Examples of the refinement model predictions can be seen in Fig.~\ref{fig_compare_olympic}. The left side shows some examples of good rink registration, while the right side shows examples of misalignments. The top row also has the segmentation overlaid on top, highlighting the slight issues with the segmentation mentioned before.

\section{Conclusion}
In this paper, we presented a novel approach to sports rink registration, by using a 3 part pipeline that is generalized to work on multiple rink types, despite only having labelled data for NHL rinks. The models are able to learn how to process different rink types and overcome a lack of labelled training data. This is done by using domain adaptation and augmentation techniques in the segmentation module, along with synthetic data and self-supervised methods in the homography and refinement modules. By doing this, we do not need additional labelled training data for other rink types, thus greatly saving annotation time and effort. This also produces a single model capable of working on multiple rink types. 

Results show that the current pipeline is competitive with results obtained by supervised NHL trained models, while also having the ability to estimate homography for non-NHL rink types as well, demonstrating great potential. Some improvements in segmentation and handling of segmentation inaccuracies can be made to further improve the robustness and accuracy of the pipeline.

\section{Acknowledgment}

This work was supported by Stathletes through the Mitacs Accelerate Program and the Natural Sciences and Engineering Research Council of Canada (NSERC).

\bibliographystyle{ieeetr}
\bibliography{egbib}

\begin{thebibliography}{10}

\bibitem{Weiner_2009}
E.~Weiner, ``Not every 200 foot by 85 foot nhl rink is the same.'' \url{https://www.nhl.com/news/not-every-200-foot-by-85-foot-nhl-rink-is-the-same/c-501626}, Oct 2009.

\bibitem{shi2022self}
F.~Shi, P.~Marchwica, J.~C.~G. Higuera, M.~Jamieson, M.~Javan, and P.~Siva, ``Self-supervised shape alignment for sports field registration,'' in {\em Proceedings of the IEEE/CVF Winter Conference on Applications of Computer Vision}, pp.~287--296, 2022.

\bibitem{jiang2020optimizing}
W.~Jiang, J.~C.~G. Higuera, B.~Angles, W.~Sun, M.~Javan, and K.~M. Yi, ``Optimizing through learned errors for accurate sports field registration,'' in {\em Proceedings of the IEEE/CVF Winter Conference on Applications of Computer Vision}, pp.~201--210, 2020.

\bibitem{nie2021robust}
X.~Nie, S.~Chen, and R.~Hamid, ``A robust and efficient framework for sports-field registration,'' in {\em Proceedings of the IEEE/CVF Winter Conference on Applications of Computer Vision}, pp.~1936--1944, 2021.

\bibitem{IIHF}
``International ice hockey federation ice rink guide.'' \url{https://www.iihf.com/en/static/5890/iihf-ice-rink-guide}.

\bibitem{Formánek}
M.~Formánek, ``Nokia arena, tampere.'' \url{https://www.eurohockey.com/arena/2154-nokia-arena-tampere.html}.

\bibitem{lowe2004distinctive}
D.~G. Lowe, ``Distinctive image features from scale-invariant keypoints,'' {\em International journal of computer vision}, vol.~60, no.~2, pp.~91--110, 2004.

\bibitem{rublee2011orb}
E.~Rublee, V.~Rabaud, K.~Konolige, and G.~Bradski, ``Orb: An efficient alternative to sift or surf,'' in {\em 2011 International conference on computer vision}, pp.~2564--2571, Ieee, 2011.

\bibitem{fischler1981random}
M.~A. Fischler and R.~C. Bolles, ``Random sample consensus: a paradigm for model fitting with applications to image analysis and automated cartography,'' {\em Communications of the ACM}, vol.~24, no.~6, pp.~381--395, 1981.

\bibitem{hartley2003multiple}
R.~Hartley and A.~Zisserman, {\em Multiple view geometry in computer vision}.
\newblock Cambridge university press, 2003.

\bibitem{detone2016deep}
D.~DeTone, T.~Malisiewicz, and A.~Rabinovich, ``Deep image homography estimation,'' {\em arXiv preprint arXiv:1606.03798}, 2016.

\bibitem{zhou2019stn}
Q.~Zhou and X.~Li, ``Stn-homography: Direct estimation of homography parameters for image pairs,'' {\em Applied Sciences}, vol.~9, no.~23, p.~5187, 2019.

\bibitem{homayounfar2017sports}
N.~Homayounfar, S.~Fidler, and R.~Urtasun, ``Sports field localization via deep structured models,'' in {\em Proceedings of the IEEE Conference on Computer Vision and Pattern Recognition}, pp.~5212--5220, 2017.

\bibitem{chen2019sports}
J.~Chen and J.~J. Little, ``Sports camera calibration via synthetic data,'' in {\em Proceedings of the IEEE/CVF conference on computer vision and pattern recognition workshops}, pp.~0--0, 2019.

\bibitem{sha2020end}
L.~Sha, J.~Hobbs, P.~Felsen, X.~Wei, P.~Lucey, and S.~Ganguly, ``End-to-end camera calibration for broadcast videos,'' in {\em Proceedings of the IEEE/CVF conference on computer vision and pattern recognition}, pp.~13627--13636, 2020.

\bibitem{chu2022sports}
Y.-J. Chu, J.-W. Su, K.-W. Hsiao, C.-Y. Lien, S.-H. Fan, M.-C. Hu, R.-R. Lee, C.-Y. Yao, and H.-K. Chu, ``Sports field registration via keypoints-aware label condition,'' in {\em Proceedings of the IEEE/CVF Conference on Computer Vision and Pattern Recognition}, pp.~3523--3530, 2022.

\bibitem{long2015fully}
J.~Long, E.~Shelhamer, and T.~Darrell, ``Fully convolutional networks for semantic segmentation,'' in {\em Proceedings of the IEEE conference on computer vision and pattern recognition}, pp.~3431--3440, 2015.

\bibitem{ronneberger2015u}
O.~Ronneberger, P.~Fischer, and T.~Brox, ``U-net: Convolutional networks for biomedical image segmentation,'' in {\em Medical Image Computing and Computer-Assisted Intervention--MICCAI 2015: 18th International Conference, Munich, Germany, October 5-9, 2015, Proceedings, Part III 18}, pp.~234--241, Springer, 2015.

\bibitem{chen2017deeplab}
L.-C. Chen, G.~Papandreou, I.~Kokkinos, K.~Murphy, and A.~L. Yuille, ``Deeplab: Semantic image segmentation with deep convolutional nets, atrous convolution, and fully connected crfs,'' {\em IEEE transactions on pattern analysis and machine intelligence}, vol.~40, no.~4, pp.~834--848, 2017.

\bibitem{chen2017rethinking}
L.-C. Chen, G.~Papandreou, F.~Schroff, and H.~Adam, ``Rethinking atrous convolution for semantic image segmentation,'' {\em arXiv preprint arXiv:1706.05587}, 2017.

\bibitem{chen2018encoder}
L.-C. Chen, Y.~Zhu, G.~Papandreou, F.~Schroff, and H.~Adam, ``Encoder-decoder with atrous separable convolution for semantic image segmentation,'' in {\em Proceedings of the European conference on computer vision (ECCV)}, pp.~801--818, 2018.

\bibitem{xie2021segformer}
E.~Xie, W.~Wang, Z.~Yu, A.~Anandkumar, J.~M. Alvarez, and P.~Luo, ``Segformer: Simple and efficient design for semantic segmentation with transformers,'' {\em Advances in Neural Information Processing Systems}, vol.~34, pp.~12077--12090, 2021.

\bibitem{long2015learning}
M.~Long, Y.~Cao, J.~Wang, and M.~Jordan, ``Learning transferable features with deep adaptation networks,'' in {\em International conference on machine learning}, pp.~97--105, PMLR, 2015.

\bibitem{ganin2016domain}
Y.~Ganin, E.~Ustinova, H.~Ajakan, P.~Germain, H.~Larochelle, F.~Laviolette, M.~Marchand, and V.~Lempitsky, ``Domain-adversarial training of neural networks,'' {\em The journal of machine learning research}, vol.~17, no.~1, pp.~2096--2030, 2016.

\bibitem{tranheden2021dacs}
W.~Tranheden, V.~Olsson, J.~Pinto, and L.~Svensson, ``Dacs: Domain adaptation via cross-domain mixed sampling,'' in {\em Proceedings of the IEEE/CVF Winter Conference on Applications of Computer Vision}, pp.~1379--1389, 2021.

\bibitem{hoyer2022daformer}
L.~Hoyer, D.~Dai, and L.~Van~Gool, ``Daformer: Improving network architectures and training strategies for domain-adaptive semantic segmentation,'' in {\em Proceedings of the IEEE/CVF Conference on Computer Vision and Pattern Recognition}, pp.~9924--9935, 2022.

\bibitem{hoyer2022hrda}
L.~Hoyer, D.~Dai, and L.~Van~Gool, ``Hrda: Context-aware high-resolution domain-adaptive semantic segmentation,'' in {\em Computer Vision--ECCV 2022: 17th European Conference, Tel Aviv, Israel, October 23--27, 2022, Proceedings, Part XXX}, pp.~372--391, Springer, 2022.

\bibitem{hoyer2023mic}
L.~Hoyer, D.~Dai, H.~Wang, and L.~Van~Gool, ``Mic: Masked image consistency for context-enhanced domain adaptation,'' in {\em Proceedings of the IEEE/CVF Conference on Computer Vision and Pattern Recognition}, pp.~11721--11732, 2023.

\bibitem{lee2019image}
M.~C. Lee, O.~Oktay, A.~Schuh, M.~Schaap, and B.~Glocker, ``Image-and-spatial transformer networks for structure-guided image registration,'' in {\em Medical Image Computing and Computer Assisted Intervention--MICCAI 2019: 22nd International Conference, Shenzhen, China, October 13--17, 2019, Proceedings, Part II 22}, pp.~337--345, Springer, 2019.

\bibitem{jaderberg2015spatial}
M.~Jaderberg, K.~Simonyan, A.~Zisserman, {\em et~al.}, ``Spatial transformer networks,'' {\em Advances in neural information processing systems}, vol.~28, 2015.

\bibitem{sinclair2022atlas}
M.~Sinclair, A.~Schuh, K.~Hahn, K.~Petersen, Y.~Bai, J.~Batten, M.~Schaap, and B.~Glocker, ``Atlas-istn: joint segmentation, registration and atlas construction with image-and-spatial transformer networks,'' {\em Medical Image Analysis}, vol.~78, p.~102383, 2022.

\bibitem{zhang2021high}
N.~Zhang and E.~Izquierdo, ``A high accuracy camera calibration method for sport videos,'' in {\em 2021 International Conference on Visual Communications and Image Processing (VCIP)}, pp.~1--5, IEEE, 2021.

\bibitem{ghiasi2021simple}
G.~Ghiasi, Y.~Cui, A.~Srinivas, R.~Qian, T.-Y. Lin, E.~D. Cubuk, Q.~V. Le, and B.~Zoph, ``Simple copy-paste is a strong data augmentation method for instance segmentation,'' in {\em Proceedings of the IEEE/CVF conference on computer vision and pattern recognition}, pp.~2918--2928, 2021.

\bibitem{tarvainen2017mean}
A.~Tarvainen and H.~Valpola, ``Mean teachers are better role models: Weight-averaged consistency targets improve semi-supervised deep learning results,'' {\em Advances in neural information processing systems}, vol.~30, 2017.

\bibitem{Iakubovskii:2019}
P.~Iakubovskii, ``Segmentation models pytorch.'' \url{https://github.com/qubvel/segmentation_models.pytorch}, 2019.

\bibitem{lin2017focal}
T.-Y. Lin, P.~Goyal, R.~Girshick, K.~He, and P.~Doll{\'a}r, ``Focal loss for dense object detection,'' in {\em Proceedings of the IEEE international conference on computer vision}, pp.~2980--2988, 2017.

\end{thebibliography}

\end{document}